\documentclass{amsart}
\usepackage[utf8]{inputenc}
\usepackage{amssymb,amsmath,amsfonts,amsthm}
\usepackage[foot]{amsaddr}
\usepackage{geometry}
\usepackage{setspace,multicol,multirow,enumerate,textcomp}
\usepackage{graphicx}
\usepackage{verbatim}
\usepackage{url}
\usepackage{hyperref}
\usepackage[dvipsnames]{xcolor}

\title{A note on the recent empirical evaluation of RBG and Ludii}
\author{Jakub Kowalski}
\email{jakub.kowalski@cs.uni.wroc.pl}
\author{Maksymilian Mika}
\email{mika.maksymilian@gmail.com}
\author{Jakub Sutowicz}
\email{jakubsutowicz@gmail.com}
\author{Marek Szyku{\l}a}
\email{msz@cs.uni.wroc.pl}
\address{Institute of Computer Science, University of Wroc{\l}aw, Wroc{\l}aw, Poland}
\begin{document}
\begin{abstract}
We present an experimental comparison of the efficiency of three General Game Playing systems in their current versions: Regular Boardgames (RBG 1.0), Ludii~0.3.0, and a Game Description Language (GDL) propnet.
We show that in general, RBG is currently the fastest GGP system.
For example, for chess, we demonstrate that RBG is about 37 times faster than Ludii, and Ludii is about 3 times slower than a GDL propnet.
Referring to the recent comparison [\emph{An Empirical Evaluation of Two General Game Systems: Ludii and RBG}, CoG 2019], we show evidences that the benchmark presented there contains a number of significant flaws that lead to wrong conclusions.
\end{abstract}
\maketitle

\section{Introduction}

In this note, we are referring to three General Game Playing (GGP) standards described in the literature: Regular Boardgames (RBG) \cite{Kowalski2019RegularBoardgames} -- a novel formalism describing via regular expressions deterministic perfect information games played on arbitrary graph boards; Ludii \cite{piette2019ludii} -- created mainly for modelling traditional strategy games and charting their historical development, a descendant of Ludi \cite{Browne10}; and the classic Stanford's GGP \cite{Genesereth2005General} based on Game Description Language (GDL) \cite{Love2008General}, which uses first-order logic to encode transition systems of all deterministic perfect information games in a very knowledge-free way. The most efficient implementations of GDL reasoners are based on Propositional Networks (propnets) \cite{Sironi2016Optimizing}.

The benchmark recently published in~\cite{piette2019empirical} contains a number of significant flaws skewing the results in favour of Ludii.
Despite our requests, the authors did not share the Ludii version used for that experiment nor they turned into discussion about its correctness.
Nevertheless, an analysis of the results and of the public Ludii versions provides strong evidence for the errors.

In this note, we show a fair benchmark, demonstrating that RBG in its first version is generally the fastest GGP system.
Then, through a detailed analysis, we point out the errors of the previous benchmark~\cite{piette2019empirical}, explaining the differences in the results.
Finally, we describe how to independently reproduce our experiment.

The most important error is that the comparison was made between games with significantly different rules, i.e., Ludii under the same name had much simpler and easier for computing rules (e.g., omitting special cases, splitting complicated moves into multiple turns).
An extreme example how the flaws affected the conclusions is the case of chess.
The authors presented that Ludii is apparently 12,000 times faster than reasoning in GDL.
However, a proper benchmark reveals that the current version of Ludii is 3 times slower than a GDL propnet for chess, apart from the existing bugs in the Ludii's chess rules.

\section{Our benchmark}\label{sec:our_benchmark}

Table~\ref{tab:experiments-mc} presents a benchmark, comparing the performance between the same games (with a few marked exceptions, with rather minor influence on the performance, explained later).
By the same games we mean the games with isomorphic game trees induced by the rules.
This assumption is required to ensure fairness when comparing efficiency of game reasoning for games encoded via different formalisms.

The computations were made on a single core of Intel(R) Core(TM) i7-4790 @3.60GHz of a computer with 16GB RAM.
The gcc version was \texttt{gcc (Ubuntu 7.4.0-1ubuntu1~18.04.1) 7.4.0}, and the Java version was \texttt{Java(TM) SE Runtime Environment (build 12.0.2+10)}.

Each test of RBG compiler took at least 10~minutes, not counting preprocessing.
Each GDL propnet result is the average time of 10 runs lasting 1 minute, not counting preprocessing (averaging is a proper practice here, because of non-deterministic propnet construction, cf.\ \cite{Sironi2016Optimizing}).
Each Ludii test was performed via the dedicated menu option.
Although theoretically Ludii~0.3.0 has a benchmark option callable from command-line, it is left completely undocumented, hence, currently, we have treat the menu option as the only official possibility admitted by the authors.

In all cases, the old version RBG 1.0 of the reasoning engine was used. Some additional games were added in RBG 1.0.1 to match their Ludii equivalents (see Subsection~\ref{subsec:games_in_rbg}).

\begin{table*}[!ht]\renewcommand{\arraystretch}{1.2}
\newcommand{\rowt}[1]{\multirow{2}{*}{#1}}
\newcommand{\rowtt}[1]{\multirow{3}{*}{#1}}
\newcommand{\col}[1]{\multicolumn{1}{c|}{#1}}
\newcommand{\colL}[1]{\multicolumn{1}{c||}{#1}}
\newcommand{\colc}[1]{\multicolumn{1}{|c|}{#1}}
\newcommand{\colcL}[1]{\multicolumn{1}{|c||}{#1}}
\newcommand{\colt}[1]{\multicolumn{2}{c|}{#1}}
\newcommand{\coltL}[1]{\multicolumn{2}{c||}{#1}}
\newcommand{\coltt}[1]{\multicolumn{3}{c|}{#1}}
\newcommand{\colttt}[1]{\multicolumn{4}{c|}{#1}}
\newcommand{\fa}{*}
\newcommand{\nda}{\hphantom{\dag}}
\newcommand{\ns}{\hphantom{*}}
\caption{Efficiency of reasoning in RBG, Ludii, and GDL propnet reasoner for the \textbf{flat Monte Carlo} test.
The values are the \textbf{numbers of playouts per second}.}\label{tab:experiments-mc}
\begin{center}\begin{tabular}{|l|r|r|r|r|r|r|}\hline
{\bf Game}            &{\bf RBG 1.0/1.0.1}&{\bf Ludii 0.3.0}&{\bf GDL propnet}\\\hline
Amazons               &             569\ns&   \emph{n/a}\nda&               4 \\\hline
Amazons-split         &           8,798\ns&        3,859\nda&             365 \\\hline
Arimaa                &            0.14*  &   \emph{n/a}\nda&      \emph{n/a} \\\hline
Arimaa-split          &             666*  &          446\dag&      \emph{n/a} \\\hline
Breakthrough          &          19,916\ns&        3,546\nda&           2,735 \\\hline
Chess-fifty move      &             523*  &           14\dag&              45 \\\hline
Connect-4             &         190,171\ns&       63,427\nda&          45,894 \\\hline
English draughts-split&          23,361*  &        7,111\dag&           3,466 \\\hline
Gomoku-free style     &           2,430*  &       26,878\nda&      \emph{n/a} \\\hline
Hex                   &           6,794\ns&       10,625\nda&      \emph{n/a} \\\hline
Reversi               &           8,682\ns&        1,312\nda&             373 \\\hline
Skirmish              &            3,989* &          636\dag&             237 \\\hline
The mill game         &          10,102*  &        2,467\dag&      \emph{n/a} \\\hline
Tic-tac-toe           &         526,930\ns&      422,836\nda&         104,500 \\\hline
\end{tabular}\end{center}
\begin{flushright}
*\ This game code was not originally available in RBG 1.0 and was added later.\\
\dag\ The rules in Ludii differ from the others.

\end{flushright}
\end{table*}

\subsection{Games in Ludii 0.3.0}\label{subsec:games_in_ludii}

Unfortunately, Ludii is a closed-source system, which limits debug possibilities for game rules.
In multiple cases, the game rules provided with Ludii 0.3.0 differ from the orthodox ones (official rules, which are encoded in RBG and GDL).
Of course, in many cases these are just regular bugs little affecting the efficiency, so mostly harmless for an efficiency comparison.
But there are also significant simplifications (as move splitting) and rule omissions, which make the computation easier.

We describe the details of the game rules occurring in both benchmarks, Table~\ref{tab:experiments-mc} and \cite{piette2019empirical}.
\begin{enumerate}
\item Amazons: 
\begin{itemize}
\item The only available version of amazons is with split movements.
In the split version, moving a queen and firing an arrow is performed in two consecutive turns, which implies that the branching factor (number of legal moves) is much smaller; this property is crucial for MC-based tests, where in each state all legal moves are computed (see \cite{kloetzer2007monte} for a comparison using MCTS). To illustrate this issue, we also present both RBG and GDL results for the orthodox (non-split) version.
\end{itemize}

\item Arimaa:
\begin{itemize}
\item Ludii contains only a split version, where each step is performed in a separate turn. Additionally, finishing a full turn is a move itself, unless there are no more steps.
\item The Ludii's version violates the constraint that a turn must make the position changed.
\item The pull action is additionally split into two moves, in contrast with the push action, which is not.
\item The traps always kill a piece immediately, regardless of the presence of a protecting neighbor.
Because of the previous point, it is also not possible to pull a piece while going into a trap.
\item The rabbit moves only forwards (does not walk horizontally).
\item There is also no ending condition in the rules other than a win. Therefore, Ludii uses its (undocumented and unknown) internal limit.
\end{itemize}

\item Chess:
\begin{itemize}
\item En-passant does not work correctly.
\item There is a number of errors related to double pawn move (e.g.\ the skipped square is impassible by sliding pieces).
\item Promotion is split into two moves.
\item It implements the fifty-move rule, which disagrees with common GDL implementations with 200 turn limit.
\end{itemize}

\item Chinese checkers:
\begin{itemize}
\item The version in Ludii is a split, where each jump is performed in a separate turn, and ending a sequence of jumps is a move itself.
\item The above fact makes this game variant absurd from the playing perspective. Indeed, a player can force a draw at almost any moment by jumping over a single piece there and back, until the internal turn limit passes.
\item There is no rule about non-leaving the goal area once reached.
\item There is no rule about not having a legal move.
\end{itemize}

\item Double chess (skirmish):
\begin{itemize}
\item It is similar to skirmish (see below). The difference is just the larger board and that promotion is allowed only to queen.
\end{itemize}

\item English draughts:
\begin{itemize}
\item The version in Ludii is a split, where each single capture is performed as a separate move.
\item Within one capture sequence (one full move), it is possible to jump with different pieces. However, a capturing sequence ends when the last used piece has no more captures. Therefore, for instance, if each of pieces A and B can perform two captures, one can jump with piece A, then jump with piece B, and then continue capturing with piece A.
\item There is no orthodox ending condition. The game ends when a player has no legal move or the internal turn limit passes.
\end{itemize} 

\item International draughts:
\begin{itemize}
\item Similarly to English draughts, the version in Ludii is a split.
\item It inherits the same bugs as English draughts.
\item The rules that a player must capture the largest possible number of pieces, and the pieces are removed from the board only after the full capturing move, are omitted.
\item Promotion occurs right after arriving at the last rank as in English draughts, in contrast to the rule that it should occur only when a sequence of captures ends there.
\end{itemize}

\item Skirmish:
The version of skirmish chess variant used in Ludii does not match any available GDL skirmish. The GDL/RBG embeds standard rules of chess except the king becomes a regular piece.
The Ludii's version is simpler due to the following:
\begin{itemize}
\item There is no castling nor en-passant.
\item Pawn promotion is split into two moves.
\item Also, there is no turn limit in the game rules, thus the internal limit must be used.
\end{itemize}

\item The mill game:
\begin{itemize}
\item This is a split version. Placing or moving a piece is performed in a separate turn than capturing.
\item The Ludii version omits the rule that an opponent's piece cannot be captured if it is in a mill, unless all his pieces are in the mill.
\item There is no flying phase when the player has only 3 pieces left.
\end{itemize}
\end{enumerate}

The remaining compared games from 0.3.0 version: breakthrough, connect-4, gomoku-free style, hex, reversi, and tic-tac-toe, are encoded, according to our knowledge, without any errors.

\subsection{Games in Regular Boardgames 1.0.1}\label{subsec:games_in_rbg}

Since the Ludii authors invented their own rules for many games, there were no equivalent versions available in RBG 1.0.
However, we made an effort to encode some of these variants (not copying the bugs, however).
They have been included to the benchmark for a rough comparison.
Due to the corruption, uncertainty, and the lack of debug possibilities of the rules of games present in Ludii, an exact comparison for such games is impossible or would require too much effort.

The new games are a part of RBG 1.0.1 and these cases are marked with star in Table~\ref{tab:experiments-mc}; the other game codes are exactly those from RBG 1.0.
The names of the games from our comparison directly match the filenames in the RBG repository, i.e., in the order from Table~\ref{tab:experiments-mc}: \texttt{amazons.rbg}, \texttt{amazons-split.rbg}, \texttt{arimaa-split.rbg}, \texttt{breakthrough.rbg}, \texttt{chess-fiftyMove}, \texttt{connect4.rbg}, \texttt{englishDraughts-split.rbg}, \texttt{gomoku-freeStyle.rbg}, \texttt{hex.rbg}, \texttt{reversi.rbg}, \texttt{skirmish.rbg}, \texttt{theMillGame-split.rbg}, \texttt{ticTacToe.rbg}.

\subsection{Games in GDL}

Whenever an equivalent GDL code is available, we also compared with the result from a GDL propnet (described in \cite{Sironi2016Optimizing}).

We additionally modified the existing GDL code for chess (the fastest version, which originally implements 200 turn limit) to match with the fifty-move rule.
Also, we fixed the GDL orthodox version of amazons for the terminal condition, which was wrong.

The GDL codes used in the experiment were:
amazons -- \texttt{amazons\_fixed.kif} from \cite{DresdenGPPServer},
amazons-split -- \texttt{amazons\_10x10.kif} from \cite{Schreiber2016Games},
breakthrough -- \texttt{breakthrough.kif} from \cite{Schreiber2016Games},
chess-fifty move -- \texttt{chessFiftyMove} from \cite{DresdenGPPServer},
connect-4 -- \texttt{connectFour.kif} from \cite{Schreiber2016Games},
English draughts-split -- \texttt{englishDraughts.kif},
reversi -- \texttt{reversi.kif} from \cite{Schreiber2016Games},
skirmish -- \texttt{skirmishZeroSum.kif} from \cite{Schreiber2016Games},
tic-tac-toe -- \texttt{ticTacToe.kif} from \cite{Schreiber2016Games}.

\section{Comparison with the other benchmark}

We provide an explanation why the results from~\cite{piette2019empirical} are much different than those in our proper experiment, leading to an opposite conclusion.
We show that that benchmark contains significant flaws of different kind.
The issues make an impression that the Ludii reasoning engine is generally faster than the others.

\begin{enumerate}
\item The most significant issue is that most of the compared games do not have same rules in all the three GGP systems.
Only 5 out of 14 games were correctly matched with their equivalents.
The differences in the other games are significant, often simplifying the computation in favor for Ludii.
We give a detailed analysis below.

\item In two cases, connect-4 and reversi, the RBG result was miscalculated.
According to our benchmark, the reported values are respectively at least about 2 and 4 times smaller than it should be, and they fall just below the Ludii results for these games.
The differences certainly are much too large to be explained by hardware differences, which is of similar performance.
It is difficult to say whether the other RBG results were computed correctly.
In general, in our test we obtained larger values for RBG and smaller values for Ludii with respect to those from~\cite{piette2019ludii}.

\item The results for GDL were obtained on a different hardware than those for RBG and Ludii.
Indeed, exactly the same GDL results were reported before in~\cite{piette2019ludii}, where they were obtained with a slower processor.
Furthermore, the result for chess was produced using the GGP-Prover instead of a propnet, despite the fact that there was available an efficient chess implementation working well under a propnet (\texttt{chess\_200.kif} from \cite{Schreiber2016Games}).
\end{enumerate}

\subsection{Differences in game rules}

The only fully correctly matched games between RBG and Ludii were breakthrough, connect-4, hex, reversi, and tic-tac-toe.
The same issue concern matching the GDL versions, as they correspond with RBG.

The authors did not share their Ludii version used for that experiment, thus it is impossible to check directly what exactly game rules were used there.
However, there are strong indirect evidences for inaccuracies.
\begin{itemize}
\item First, correct rules of the concerned games in a proper variant, fully corresponding to those in RBG 1.0, are not present in any of the published Ludii versions.
Thus, it would be unbelievable that they existed in the past and were used for the benchmark.
\item Second, the current Ludii 0.3.0 version exhibits similar performance compared to the results from~\cite{piette2019empirical}.
This shows that in terms of efficiency, there are only minor differences between these versions, which are not meaningful for the conclusions.
There are two exceptions, chess and the mill game, for which the public version is significantly slower; this suggests that a much simpler ruleset must had been used.
\end{itemize}

Note that in our experiment (Table~\ref{tab:experiments-mc}), we have used the same version RBG~1.0 as in~\cite{piette2019ludii}.
A few new game variants were added, because otherwise we could not provide any meaningful comparison with Ludii for them.
They roughly demonstrate that the situation in that benchmark would be different if the games were more correctly matched.
We have also found a few minor mistakes in the game codes of RBG 1.0, which were fixed.
Nevertheless, they had unnoticeable impact on performance, except a small difference for gomoku and the mill game.

\begin{enumerate}
\item Amazons, arimaa, Chinese checkers, English draughts, international draughts, and the mill game:
First, in Ludii, these were split versions, which were compared with orthodox rules versions in RBG.
The importance of this mismatching is clearly visible in our benchmark (see Table~\ref{tab:experiments-mc}).
We included both orthodox and split versions of amazons and arimaa in RBG, which use equivalent rules up to the split moves.
The split versions in Ludii indeed work faster than the orthodox versions in RBG, but under the proper matching, it is the opposite.
It should not be surprising, since the point of a MC test is to compute all legal moves for every visited game state.
In the extreme case of arimaa, splitting reduces branching factor from over 200,000 (the pure flat-MC test does not merge distinct legal moves leading to the same game state) to about 20 (with increased number of turns up to 4); however, the paper reported its 6,490 speedup over RBG without any meaningful analysis for possible reasons.
Furthermore, amazons-split code, which was available in RBG 1.0, was ignored.

The second reason is that, except amazons, these games had incorrect and simplified rules, as their corresponding versions in the public releases have.
The simplification is especially influential in the case of arimaa and international draughts, where Ludii's versions omit costly parts of rules (e.g., preventing from repeating the position, restricting captured opponent's pawns to those out of mill).

\item Chess and double chess:
Most likely the used version was a simpler variant of chess rules.
This conclusion is supported by the results, as the reported performance of chess is more close to that of skirmish in our benchmark.
Furthermore, this kind of simplification seems to be a common practice, as for double chess Ludii contains only the skirmish version; despite that, in the paper, it was compared with orthodox double chess in RBG.

\item Gomoku: The Ludii's version was the \emph{free style gomoku}, in contrast with the \emph{gomoku standard} in RBG and GDL. Gomoku standard was added to Ludii in~0.3.0. Indeed, in this particular case this was only a minor difference.
\end{enumerate}

\section{Reproduction of the results}

\subsection{RBG}

While the procedure below may look a bit complicated, it is to ensure that the same version RBG 1.0 is used, with the exception of the added games.

\begin{enumerate}
\item The RBG~1.0 version is available at~\cite{RBGsource100}.
\item To get the games marked with star in Table~\ref{tab:experiments-mc}, get RBG~1.0.1 from \cite{RBGsource101} and extract these games into \texttt{games} directory.
\item Run \texttt{make.sh} from \texttt{scripts}.
\item A single test can be run from \texttt{rbg2cpp} directory by:
\begin{verbatim}
make simulate_[game] SIMULATIONS=[playouts]
\end{verbatim}
where \texttt{[game]} is a game name from \texttt{games} directory.
The number of playouts should be large enough to process for at least a few seconds.
For example, let:
\begin{verbatim}
make simulate_chess-fiftyMove SIMULATIONS=10000
\end{verbatim}
In the output, it can be found the number of playouts performed in specified time in milliseconds, which is the total test time without preprocessing, e.g, \texttt{performed 10000 plays in 19038 ms.}
\end{enumerate}

\subsection{Ludii}

To test Ludii, download it from \cite{LudiiPortal}.
Having installed Java, run the Ludii player assigned to one processor core on an idle system to ensure than only one is used and to prevent migration, which gives a more stable computation; this is optional, since the results do not differ much.
On a linux system, let:
\begin{verbatim}
taskset -c 0 java -jar Ludii-0.3.0.jar
\end{verbatim}
To test a game, load it first with \texttt{File/Load~Game} and then use \texttt{Game/Time~Random~Playouts}.

\section{Conclusions}

We have performed a fair benchmark of the efficiency of three GGP systems in their current versions available at the time of writing this note.

The RBG dominates in all except two games: gomoku and hex, for which Ludii is faster.
Taking into account tic-tac-toe, where both systems have similar results, it is visible that Ludii is well optimized for this kind of games, where legal moves are actions filling one empty square on the board.
For several games in our benchmark, the version in Ludii does not encode proper rules, which also can simplify computation. E.g., in arimaa, there is no checking whether the position is the same, and improper movement of rabbits and trap logic make random playouts shorter in average.

Concerning the benchmark from~\cite{piette2019ludii}, there are many arguments against its correctness, and if done properly, the conclusions would be different.
Indeed, according to our benchmark, even if the unpublished Ludii version was, supposedly, twice faster than the public versions, that does not change the overall situation.

\section*{Acknowledgments}

We thank Chiara F.\ Sironi for sharing the GDL propnet code and for helping with using it.

\bibliographystyle{plain}

\end{document}